\documentclass[runningheads]{llncs}

\usepackage{accv}

\usepackage{booktabs}
\usepackage{tabularx}
\usepackage{array}
\usepackage{makecell}
\usepackage{booktabs}
\usepackage{xltabular}
\usepackage{array}
\usepackage{multirow}
\usepackage{caption}
\usepackage{makecell}
\usepackage{pdflscape} 
\usepackage{adjustbox}
\usepackage{accvabbrv}

\usepackage{graphicx}
\usepackage{float}
\usepackage[section]{placeins}
\usepackage{booktabs}
\usepackage{tabularx}
\usepackage{adjustbox}

\usepackage{graphicx}
\usepackage{booktabs}
\usepackage{wrapfig}

\usepackage[accsupp]{axessibility}  

\usepackage{tikz}
\usepackage{xcolor}
\usepackage{graphicx}
\usepackage{placeins}
\usetikzlibrary{arrows.meta,calc,positioning,shapes.geometric,fit,backgrounds}

\definecolor{vlmBlue}{HTML}{173B8F}
\definecolor{vlmRed}{HTML}{C51F24}
\definecolor{vlmGreen}{HTML}{087A32}
\definecolor{vlmBrown}{HTML}{8A630B}
\definecolor{vlmInk}{HTML}{1B1F23}
\definecolor{vlmGray}{HTML}{5F6770}
\definecolor{vlmLightBlue}{HTML}{EEF4FF}
\definecolor{vlmLightRed}{HTML}{FFF0F0}
\definecolor{vlmLightGreen}{HTML}{EFFAF3}
\definecolor{vlmLightBrown}{HTML}{FFF8EA}

\tikzset{
  >={Latex[length=2.2mm,width=1.8mm]},
  probeSector/.style={line width=0.7pt, fill opacity=0.50},
  probeTitle/.style={font=\sffamily\bfseries\large, align=center, text=vlmInk},
  probeText/.style={font=\sffamily\small, align=center, text=vlmInk},
  probePill/.style={draw=vlmGray!60, rounded corners=2pt, fill=white, font=\sffamily\scriptsize\bfseries, align=center, minimum height=0.34cm},
  panel/.style={rounded corners=3pt, line width=0.75pt, fill=white},
  panelTitle/.style={font=\sffamily\bfseries\large, anchor=west, align=left},
  subpanel/.style={rounded corners=3pt, line width=0.55pt, fill=white, align=center, font=\sffamily\scriptsize},
  tinyLabel/.style={font=\sffamily\tiny, align=center},
  smallLabel/.style={font=\sffamily\scriptsize, align=center},
  flowText/.style={font=\sffamily\scriptsize, align=center, text=vlmInk},
  badge/.style={circle, fill=vlmBlue, text=white, font=\sffamily\bfseries\small, inner sep=1.1pt, minimum size=0.34cm},
  redbadge/.style={badge, fill=vlmRed},
  greenbadge/.style={badge, fill=vlmGreen},
  brownbadge/.style={badge, fill=vlmBrown},
  arrowBlue/.style={->, very thick, draw=vlmBlue},
  arrowRed/.style={->, very thick, draw=vlmRed},
  arrowGreen/.style={->, very thick, draw=vlmGreen},
  arrowPurple/.style={->, very thick, draw=purple!70!black},
  dArrow/.style={->, thick, dashed, draw=#1}
}

\newcolumntype{L}[1]{>{\raggedright\arraybackslash}p{#1}}
\newcolumntype{C}[1]{>{\centering\arraybackslash}p{#1}}

\usepackage{hyperref}

\usepackage{orcidlink}

\begin{document}

\title{\textit{Do VLMs Really Look at the Road?} \\ A Diagnostic Evaluation for Autonomous Driving} 

\titlerunning{\textit{Do VLMs Really Look at the Road?}}

\author{Shankar Gangisetty$^*$\orcidlink{0000-0003-4448-5794} \and
Muskan Singh$^*$\orcidlink{0009-0002-3766-7477}  \and
C.V. Jawahar\orcidlink{0000-0001-6767-7057}}
\authorrunning{S.~Gangisetty et al.}

\institute{International Institute of Information Technology, Hyderabad, India \\
}

\maketitle

\begin{abstract}
Vision-language models (VLMs) are increasingly being evaluated for autonomous driving through question-answering benchmarks, yet high benchmark accuracy does not necessarily reflect visually grounded reasoning. We present a diagnostic evaluation framework that systematically probes whether VLMs base safety-critical driving decisions on visual evidence or instead rely on language priors, answer biases, and dataset regularities. Our benchmark comprises two task categories, Adverse Weather and Junctions \& Intersections, with a nuScenes spatial reasoning extension, and evaluates six open-source VLMs along four diagnostic protocols: image versus no-image evaluation, counterfactual weather sensitivity, linguistic robustness under paraphrasing and negation, and spatial reasoning. All ground-truth labels are derived programmatically from BDD100K and DriveLM annotations, enabling reproducible evaluation without manual labeling. Across all models, we find that competitive overall accuracy conceals systematic failure modes, including strong affirmation bias, limited sensitivity to counterfactual scene changes, near-chance performance on negated questions despite high paraphrase consistency, and confident driving decisions even without visual input. These results demonstrate that current open-source VLMs remain insufficiently grounded for safety-critical autonomous driving and underscore the need for diagnostic evaluation beyond aggregate accuracy. The code will be made publically available.

\end{abstract}

\section{Introduction}
\label{sec:intro}

Vision-language models (VLMs) are increasingly being explored as interpretable reasoning modules for autonomous driving, enabling systems to answer natural-language questions about road scenes, explain driving decisions, and support interactive perception, prediction, and planning. Recent benchmarks such as DriveLM~\cite{drivelm2024} and DriveBench~\cite{xie2025drivebench} demonstrate the potential of language-guided interfaces for evaluating and interacting with driving systems. Beyond improving interpretability, such interfaces provide a promising mechanism for inspecting model decisions in safety-critical scenarios, where understanding \emph{why} a decision is made is often as important as the decision itself.


\begin{wrapfigure}{r}{0.48\textwidth}
    \centering
    \includegraphics[width=0.46\textwidth]{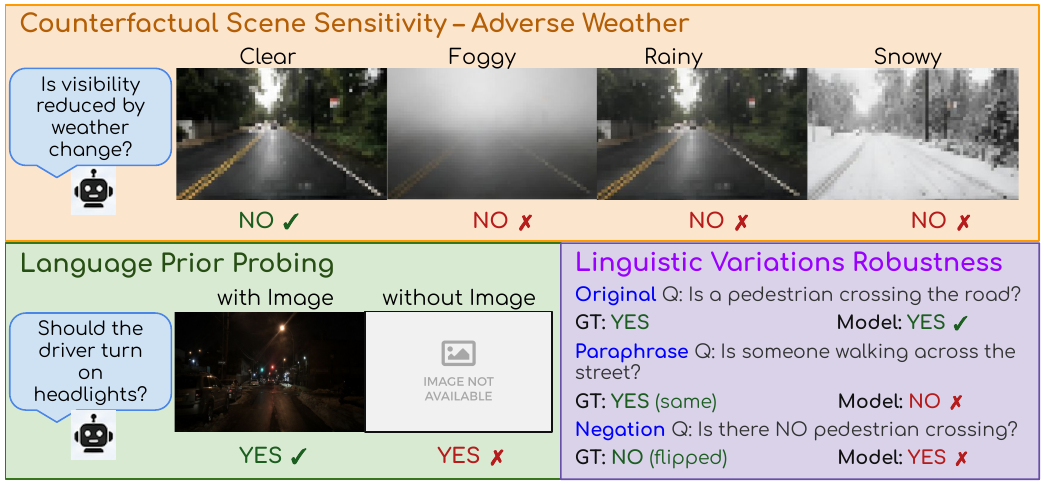}
    \caption{\textbf{Motivation for diagnostic evaluation of driving VLMs.} Existing driving benchmarks may report high accuracy even when VLMs are not visually grounded. Our framework probes three complementary failure modes: (top) counterfactual scene sensitivity under controlled weather changes, (bottom left) reliance on language priors through image versus no-image evaluation, and (bottom right) robustness to linguistic variations via paraphrases and negation. Together, these diagnostics expose failure modes that are hidden by aggregate benchmark accuracy.}
    \label{fig:teaser}
\end{wrapfigure}

Despite this promise, strong performance on driving benchmarks does not necessarily imply that a model is reasoning from visual evidence. Recent studies have shown that VLMs can fail surprisingly simple visual reasoning tasks despite competitive benchmark accuracy~\cite{rahmanzadehgervi2024blind}. Other work attributes this behavior to modality imbalance, where models rely disproportionately on language priors instead of image content~\cite{deng2025blindfaith,sim2025modalitycollapse}. In autonomous driving, such reliance is particularly concerning: a model may recommend braking, yielding, or switching on headlights because these actions are statistically associated with the question rather than because it has correctly recognized pedestrians, traffic signals, weather conditions, or other scene-specific cues. Consistent with this observation, DriveBench~\cite{xie2025drivebench} reports that modern VLMs continue to produce plausible driving responses even when visual inputs are degraded or entirely absent. These findings suggest that aggregate accuracy alone is insufficient to assess the reliability of VLMs for safety-critical driving (Fig.~\ref{fig:teaser}).

Reliable driving assistants must also remain robust to natural language variation. Driving questions frequently involve negation and polarity, for example whether a pedestrian is \textit{not} present, headlights are \textit{not} required, or a junction is \textit{unsafe} to enter. Prior work shows that VLMs perform only at or near chance on visually grounded negation tasks~\cite{alhamoud2025negbench,zhang2025negvqa}, while exhibiting an affirmation bias, i.e., predicting the queried object or condition even when explicitly negated~\cite{kang2025notdetect}. In autonomous driving, such failures translate directly into safety risks: unwarranted affirmative responses may hallucinate hazards, whereas incorrect negative responses may suppress genuine warnings. Together, these observations motivate evaluation protocols that test not only perception accuracy but also visual grounding and semantic robustness.


Motivated by these limitations, we ask a different question from existing driving benchmarks:

\emph{Do current VLMs genuinely reason from visual evidence when answering safety-critical driving questions, or do they rely on language priors, answer biases, and dataset regularities?}

To answer this question, we introduce a diagnostic evaluation framework that moves beyond aggregate accuracy to probe complementary aspects of driving reliability. The framework evaluates (i) visual grounding by comparing image-conditioned predictions against a no-image baseline, (ii) robustness to counterfactual weather changes through paired clear, fog, rain, and snow scenes, (iii) linguistic robustness using paraphrased and negated question variants, and (iv) spatial reasoning in complex intersection scenarios using DriveLM/nuScenes annotations. We instantiate this framework on two benchmark categories, \textbf{Adverse Weather} and \textbf{Junctions \& Intersections}, constructed entirely through programmatic ground-truth derivation from BDD100K~\cite{yu2020bdd100k} and DriveLM/nuScenes~\cite{drivelm2024,caesar2020nuscenes}, eliminating manual annotation while ensuring reproducibility.



Our evaluation of open-source VLMs reveals that competitive benchmark accuracy often conceals systematic failure modes. Models exhibit strong affirmation and safety-prior biases, limited sensitivity to counterfactual scene changes, weak spatial reasoning in safety-critical junctions, and near-chance performance on negated questions despite remaining highly consistent under paraphrasing. These results indicate that current VLMs frequently produce plausible driving responses without consistently grounding their decisions in visual evidence. We hope this benchmark encourages more diagnostic and safety-oriented evaluation of VLMs for autonomous driving beyond conventional accuracy-based comparisons. A comparison with existing benchmarks is provided in Table~\ref{tab:related_work}; additional discussion of related work is included in the supplementary material.


Our contributions are threefold:
\begin{itemize}
    \item \textbf{We introduce a diagnostic benchmark} for autonomous driving that evaluates visual grounding, counterfactual robustness, linguistic robustness, and spatial reasoning, moving beyond aggregate accuracy as the primary evaluation criterion.
    \item \textbf{We construct two complementary benchmark categories}, Adverse Weather and Junctions \& Intersections, using fully programmatic ground-truth derivation from BDD100K and DriveLM/nuScenes, together with counterfactual weather transformations and controlled linguistic variants.
    \item \textbf{We present a comprehensive evaluation of open-source VLMs}, revealing consistent failure modes including language-prior reliance, affirmation bias, weak sensitivity to visual scene changes, limited spatial reasoning, and poor polarity-aware reasoning despite competitive benchmark accuracy.
\end{itemize}

\begin{table}[t]
\centering
\caption{Comparison of our diagnostic framework over the closest related
benchmarks.}
\label{tab:related_work}
\small
\setlength{\tabcolsep}{2.2pt}
\begin{tabular}{lcccccc}
\toprule
& \scriptsize POPE~\cite{li2023pope} & \scriptsize  NegVQA~\cite{zhang2025negvqa} & \scriptsize  NegBench~\cite{alhamoud2025negbench} & \scriptsize  DriveLM~\cite{drivelm2024} & \scriptsize  DriveBench~\cite{xie2025drivebench} & \scriptsize  \textbf{Ours} \\
\midrule
Driving         & --        & --        & --        & \checkmark & \checkmark             & \checkmark \\
No-image       & --        & --        & --        & --        & \checkmark$^{\dagger}$  & \checkmark \\
Counterfactual   & --        & --        & --        & --        & \checkmark$^{\ddagger}$ & \checkmark \\
Paraphrase  & --        & --        & --        & --        & --                      & \checkmark \\
Negation       & --        & \checkmark & \checkmark & --       & --                      & \checkmark \\
Spatial QA       & --        & --        & --        & \checkmark & --                     & \checkmark \\
\bottomrule
\end{tabular}
{\tiny $^{\dagger}$Text-only evaluation (no image input).
$^{\ddagger}$Synthetic weather corruptions, not paired scene-level counterfactuals.}
\end{table}





\section{Diagnostic Driving Evaluation Framework}
\label{sec:framework}

\begin{figure*}[tb]
  \centering
  \includegraphics[width=\textwidth]{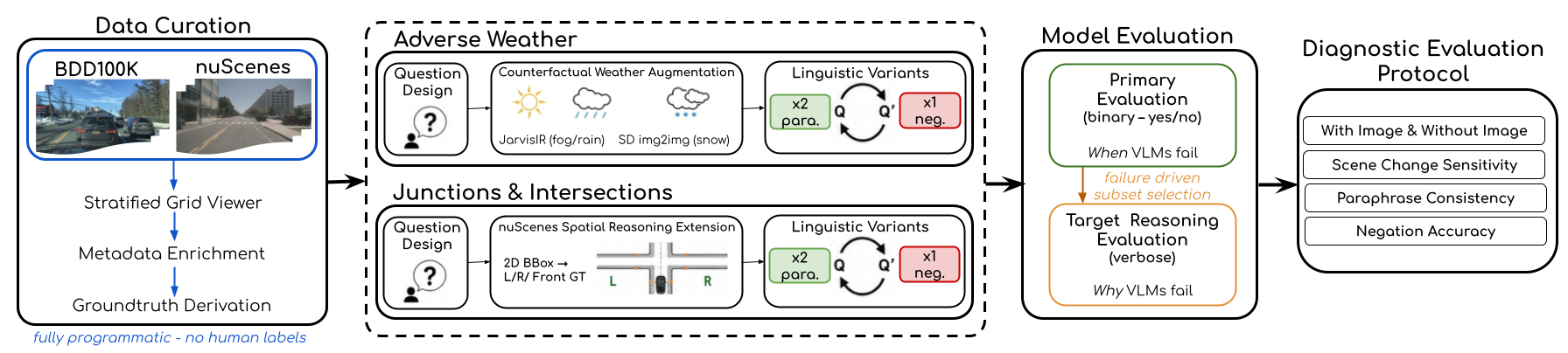}
  \caption{\textbf{Overview of the proposed diagnostic driving evaluation framework.} BDD100K and nuScenes images are curated with metadata for programmatic ground-truth derivation. The Adverse Weather task introduces weather counterfactuals (fog, rain, snow) and paraphrase/negation variants, while Junctions \& Intersections extends the benchmark with nuScenes spatial reasoning and the same linguistic variants. VLMs are first evaluated using binary yes/no prompts across all diagnostic protocols, followed by failure-driven targeted reasoning analysis using evidence-based prompts.}
  \label{fig:pipeline}
\end{figure*}

Fig.~\ref{fig:pipeline} provides an overview of the proposed diagnostic driving evaluation framework. We first describe the benchmark construction, including data curation, programmatic ground-truth derivation, and question design for the Adverse Weather and Junctions \& Intersections tasks. We then present the diagnostic evaluation protocol and metrics for assessing visual grounding, counterfactual robustness, linguistic robustness, and spatial reasoning.

\subsection{Data Curation}
The benchmark comprises 520 images in total: 320 from BDD100K (including 72 counterfactual weather images) and 200 CAM\_FRONT keyframes from DriveLM/nuScenes. This yields 26,788 binary question--answer pairs. More detailed statistics are provided in the Supplementary Material.

\noindent\textbf{BDD100K.}
The primary benchmark is constructed from the BDD100K validation split~\cite{yu2020bdd100k}. Images are manually curated using an interactive stratified grid viewer that groups candidate frames by driving condition, enabling balanced selection while avoiding visually redundant scenes. For the \textbf{Adverse Weather} category, 108 images are selected across five conditions (26 rainy, 16 foggy, 25 snowy, 17 night-time, and 24 clear-daytime scenes). For the \textbf{Junctions \& Intersections} category, 140 images are selected across five junction types (26 signalised, 28 multi-signal, 27 pedestrian-heavy, 35 unsignalised, and 24 highway-merge intersections). In total, the benchmark contains 248 uniquely curated BDD100K images with balanced coverage of diverse driving scenarios.

\noindent\textbf{nuScenes and DriveLM.}
To extend the Junctions \& Intersections benchmark beyond scene-level reasoning, we additionally sample 200 CAM\_FRONT keyframes from the DriveLM v1.1 annotations~\cite{drivelm2024} on the nuScenes dataset~\cite{caesar2020nuscenes}. The subset is stratified over intersection scenarios and pedestrian presence, while restricting evaluation to the front camera to leverage its camera-attributed 2D object annotations. These keyframes support spatial, predictive, and planning-oriented reasoning tasks that complement the BDD100K benchmark.

\subsection{Ground-Truth Derivation}

Ground truth is generated entirely through deterministic functions operating on existing dataset annotations, eliminating manual question-answer labelling. Each question is associated with a derivation function that maps annotation attributes to a binary decision label, ensuring reproducible and scalable benchmark construction.

For BDD100K, derivation functions operate on enriched image annotations including weather, time of day, scene type, traffic-light states, per-class object annotations, bounding-box geometry, and occlusion metadata. We employ a two-tier strategy: Tier-I questions are derived directly from reliable annotation fields, whereas Tier-II questions use deterministic combinations of multiple attributes when a single annotation is insufficient to support a driving decision. This enables driver-centric questions while avoiding subjective visual judgement.

For the DriveLM/nuScenes extension, ground truth is derived from camera-attributed 2D object annotations in the CAM\_FRONT view. Relative object locations are used to infer spatial relationships (e.g., left, right, or ahead of the ego vehicle), supporting deterministic construction of spatial, prediction, and planning questions for complex intersection scenarios.

\subsection{Question Design}
Questions are formulated from the \emph{driver's perspective}, requiring models to make safety-critical driving decisions rather than merely describe the scene (e.g., \emph{Should the driver turn on the headlights?''} instead of \emph{Is it raining?''}). Each question is associated with a deterministic ground-truth derivation function, and the correct answer is designed to vary across visually similar scenes, preventing models from relying solely on dataset statistics or condition labels. The benchmark comprises two complementary category tasks: \textbf{Adverse Weather} and \textbf{Junctions \& Intersections}, each constructed to probe different aspects of visually grounded driving reasoning.


\begin{enumerate}
    \item \textbf{Adverse Weather.} This task category evaluates whether models adapt driving decisions to changing weather conditions rather than relying on fixed language or safety priors.
    \begin{itemize}
        \item \textbf{Language-prior probing.} Every binary driver-centric question is evaluated under both image-conditioned and no-image settings. Comparing the two directly measures whether a model genuinely uses visual evidence or continues to produce plausible driving decisions when visual information is absent.
        \item \textbf{Counterfactual weather augmentation.} To evaluate visual grounding, each clear-weather scene is paired with counterfactual fog, rain, and snow variants generated from the same underlying image. Fog and rain are synthesized using the physics-based degradation models of JarvisIR~\cite{lin2025jarvisir}, while snow is generated using Stable Diffusion image-to-image synthesis~\cite{rombach2022ldm}. Since only the weather condition changes while scene geometry remains fixed, answer changes can be attributed directly to weather-dependent visual reasoning. 
        \item \textbf{Semantic question types.} Questions are organised into five semantic types. Together, these span progressively more challenging reasoning levels, from recognizing visual cues to making safety-critical decisions that integrate multiple scene attributes.
        \begin{itemize}
        \item \textit{Perception}: recognising traffic lights, vehicles, pedestrians, or visibility cues.
        \item \textit{Action}: driver decisions such as headlights, speed adaptation, or overtaking.
        \item \textit{Hazard}: identifying safety-critical risks arising from weather or occlusions.
        \item \textit{Scene}: understanding road, lighting, and environmental conditions.
        \item \textit{Adversarial}: combining multiple conditions or contrastive cues to probe deeper reasoning.
        \end{itemize}
        \item \textbf{Linguistic variations.} Every question is further expanded into two meaning-preserving paraphrases and one negated variant. This separates robustness to superficial rewording from the ability to correctly reason under semantic polarity reversal.
    \end{itemize}
    \item \textbf{Junctions \& Intersections.} This category task evaluates reasoning in complex intersection scenarios where safe driving depends on understanding dynamic interactions among traffic participants.
    \begin{itemize}
        \item \textbf{Language-prior probing.} As in Adverse Weather, every question is evaluated with and without the driving image, exposing whether intersection decisions remain visually grounded or are dominated by language and safety priors.
        \item \textbf{Semantic question types.} The BDD100K benchmark includes questions covering traffic-light interpretation, pedestrian awareness, hazard recognition, driver actions, and overall scene understanding using the same five semantic categories defined for Adverse Weather.
        \item \textbf{nuScenes spatial reasoning extension.}  To probe reasoning beyond scene recognition, the benchmark is extended using DriveLM/nuScenes annotations with three additional reasoning types:
        \begin{itemize}
        \item \textit{Spatial Reasoning}: relative positions of surrounding vehicles and pedestrians (left/right/front).
        \item \textit{Prediction}: anticipated behavior of surrounding traffic participants.
        \item \textit{Planning}: ego-vehicle decisions conditioned on the spatial configuration of the scene.
        
        These questions require relational reasoning over object-level annotations rather than simple object recognition, making them substantially more diagnostic of higher-level driving intelligence.
        \end{itemize}
        \item \textbf{Linguistic variations.} The same paraphrase and negation protocol is applied, enabling direct comparison of linguistic robustness across both benchmark tasks.
    \end{itemize}
\end{enumerate}

\subsection{Model Evaluation and Diagnostic Protocol}
\label{subsec:evaluation}
The proposed framework evaluates VLMs in two stages. The first measures large-scale binary decision performance across the complete benchmark, while the second investigates \textit{why} models fail through targeted reasoning analysis. Together, these stages separate quantitative performance from qualitative failure diagnosis.\\

\noindent\textbf{Primary Evaluation.}

VLMs are first evaluated on the complete benchmark using a constrained binary prompt that requires either \textbf{Yes} or \textbf{No} as the response. This stage provides a controlled and comparable evaluation across all benchmark tasks while minimizing the influence of verbose generation styles. Responses that cannot be resolved into binary decisions are recorded as unclear and excluded from accuracy computation.

Primary evaluation is performed under four diagnostic protocols designed to expose different failure modes beyond aggregate accuracy.
\begin{itemize}
    \item \textbf{with Image.} Models receive the driving image together with a binary driving question. This represents the standard evaluation setting and measures decision accuracy when visual evidence is available.
    \item \textbf{without Image.} The driving image is replaced with a blank input while the question remains unchanged. Comparing performance between with and without image quantifies the extent to which predictions depend on visual evidence rather than language or safety priors.
    \item \textbf{Counterfactual weather evaluation.} For Adverse Weather, every clear-weather image is paired with synthesized fog, rain, and snow variants while keeping the underlying scene unchanged. Differences in predictions reveal whether driving decisions appropriately respond to controlled visual changes instead of remaining invariant.
    \item \textbf{Linguistic variation evaluation.} Each question is further evaluated using two paraphrases and one negated formulation. \textit{Paraphrases} assess robustness to meaning-preserving rewording, whereas \textit{negated} questions evaluate polarity-aware reasoning under semantic reversal.
\end{itemize}
For the Junctions \& Intersections benchmark, we additionally evaluate a nuScenes spatial reasoning extension, containing spatial reasoning, prediction, and planning questions derived from DriveLM annotations. This extension examines whether models generalize beyond scene recognition to higher-level relational reasoning required for autonomous driving.\\

\noindent\textbf{Targeted Reasoning Evaluation.}

Binary predictions alone provide limited insight into the causes of model failures. We therefore perform a second evaluation stage on a failure-driven subset identified from the primary evaluation. This targeted analysis is conducted on representative VLMs, selected to represent different performance and bias characteristics.
Instead of binary responses, models are prompted to produce a binary decision followed by a short evidence-based explanation describing the visual cues supporting their answer. By comparing explanations across image-conditioned and no-image settings, as well as across counterfactual and linguistic variants, this stage reveals whether failures arise from missing visual grounding, reliance on language or safety priors, hallucinated evidence, or incorrect semantic reasoning.

Rather than relying solely on aggregate accuracy, the benchmark reports diagnostic metrics that quantify different aspects of model reliability discussed in Supplementary Material.

\section{Experimental Results}
\label{sec:results}


\subsection{Implementation Details}
 
We evaluate six representative open-source VLMs: Moondream 2B~\cite{moondream2025}, SmolVLM 2.2B~\cite{smolvlm2025}, PaliGemma 3B~\cite{paligemma2024}, Qwen2.5-VL 7B~\cite{qwen25vl2025}, InternVL3 8B~\cite{internvl32025}, LLaVA-OV 8B~\cite{llavaonevision2024}. All models are evaluated in a strictly zero-shot setting without task-specific fine-tuning,
in-context examples, or additional prompt engineering.

Inference is performed using Hugging Face \texttt{transformers} library with models loaded in \texttt{bfloat16} precision on a single NVIDIA RTX 3080 GPU. Greedy decoding is used throughout, with a maximum generation length of 50 tokens for the primary binary evaluation and 200 tokens for the targeted reasoning analysis. Binary predictions are extracted using a prefix scan for \texttt{yes}/\texttt{no}, followed by a keyword-based fallback over the first generated sentence. Responses that cannot be unambiguously mapped to either class are recorded as \texttt{unclear} and excluded from quantitative analysis. Due to restrictive safety guardrails, PaliGemma-3B and SmolVLM-2.2B are excluded from targeted reasoning analysis.

\subsection{Performance Analysis on Adverse Weather}
 
\noindent\textbf{Language-prior (with Image).}
Table~\ref{tab:aw_axis1} presents results under the image-conditioned setting. Although the strongest models achieve around 76--78\% overall accuracy, the benchmark reveals that comparable performance can arise from fundamentally different decision strategies. GT-conditioned analysis exposes pronounced response biases that are largely invisible from aggregate accuracy alone. Moondream~2B (+29.5,pp) and InternVL3~8B (+18.4,pp) consistently over-affirm, whereas SmolVLM~2.2B exhibits the opposite behaviour (-42.0,pp), defaulting to negative responses irrespective of the visual scene. Such asymmetric biases are particularly problematic for autonomous driving, where systematic false positives or false negatives can directly translate into unsafe decisions.

\begin{figure*}[tb]
  \centering
  \includegraphics[width=\textwidth]{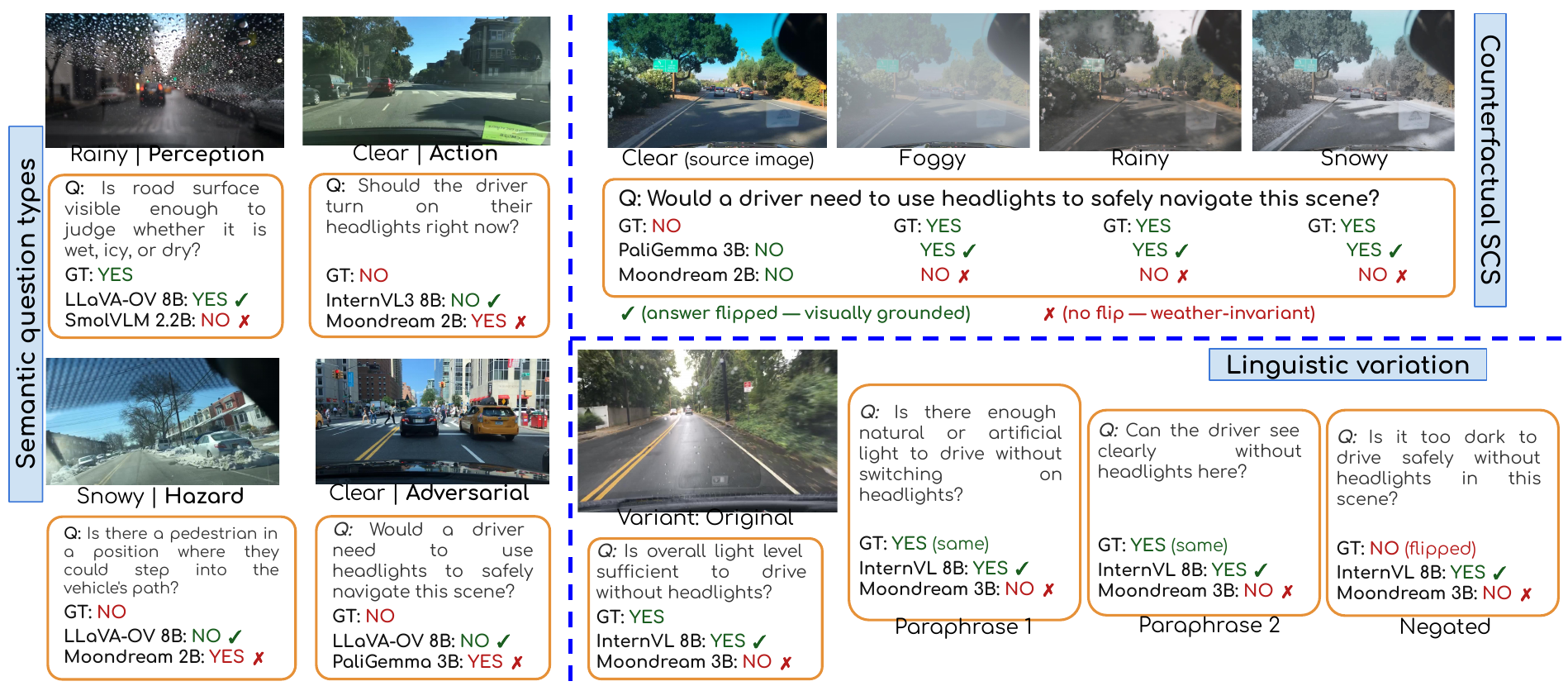}
  \caption{\textbf{Qualitative analysis of the Adverse Weather task.} \textit{Semantic question types:} best- and worst-performing models on matched image-question pairs across perception, action, hazard, and adversarial questions. \textit{Counterfactual SCS:} PaliGemma-3B adapts its prediction across clear, fog, rain, and snow, whereas Moondream-2B remains weather-invariant. \textit{Linguistic variation:} InternVL3-8B is consistent across the original question and paraphrases but fails under negation. Zoom in for details; additional details are provided in the Supplementary.}
  \label{fig:adv_weater_primary}
\end{figure*}

\noindent\textbf{Adverse weather types.} 
Performance varies substantially across weather conditions, indicating that current VLMs are not uniformly grounded in visual evidence. Surprisingly, all models perform better on night scenes than on visually simpler clear-day scenes. Since many weather-dependent driving actions (e.g., using headlights or reducing speed) are naturally affirmative at night, this pattern suggests that models exploit correlations between scene context and expected answers rather than reasoning from the observed weather itself. Moondream's sharp degradation on clear scenes (36.6\%) further illustrates its inability to suppress affirmative responses when adverse-weather actions are unnecessary, revealing reliance on learned safety priors instead of scene-specific evidence (Fig.~\ref{fig:adv_weater_primary}).


\begin{table*}[tb]
\caption{\textbf{Image-conditioned performance on the Adverse Weather task.} Results are reported in accuracy. The \textit{adverse weather type} relates to weather in the conditioned image. The affirmation gap (A. Gap) quantifies response bias, where positive (negative) values indicate over-affirmative (over-negative) behavior. \textbf{Bold} and \underline{underlined} values denote the best and worst results.}
\label{tab:aw_axis1}
\centering
\scriptsize
\setlength{\tabcolsep}{3pt}
\renewcommand{\arraystretch}{1.13}
\resizebox{\textwidth}{!}{%
\begin{tabular}{@{}lrrrr|rrrrr|rrrrr@{}}
\toprule
& \multicolumn{4}{c|}{Language-prior probing (\%)}
& \multicolumn{5}{c|}{Adverse weather type (\%)}
& \multicolumn{5}{c}{Semantic question types (\%)} \\
\cmidrule(lr){2-5} \cmidrule(lr){6-10} \cmidrule(lr){11-15}
Model
& Overall & GT=yes & GT=no & A. Gap
& Rainy & Foggy & Snowy & Night & Clear
& Perc. & Action & Hazard & Scene & Adv. \\
\midrule
Moondream 2B~\cite{moondream2025}
& \underline{60.1} & 76.0 & \underline{46.5} & $+$29.5
& 59.6 & \underline{66.1} & 67.4 & 80.2 & \underline{36.6}
& 61.9 & \underline{62.3} & \underline{43.0} & 72.1 & 62.2 \\
 
PaliGemma 3B~\cite{paligemma2024}
& 66.7 & 67.8 & 65.8 & $+$2.0
& 61.1 & 67.1 & 65.2 & \underline{75.5} & 67.6
& 65.2 & 73.7 & 66.3 & \underline{70.1} & \underline{59.1} \\
 
SmolVLM 2.2B~\cite{smolvlm2025}
& 66.7 & \underline{43.8} & \textbf{85.8} & $-$42.0
& \underline{56.1} & 70.5 & \underline{63.6} & 79.6 & 69.4
& \underline{61.1} & 78.4 & 59.2 & 68.7 & 72.7 \\
 
LLaVA-OV 8B~\cite{llavaonevision2024}
& \textbf{78.3} & 74.5 & 81.5 & $-$7.0
& 70.6 & \textbf{76.6} & \textbf{77.4} & \textbf{93.1} & 78.7
& \textbf{72.3} & 77.9 & \textbf{74.5} & \textbf{92.8} & \textbf{79.7} \\
 
InternVL3 8B~\cite{internvl32025}
& 75.8 & \textbf{85.8} & 67.4 & $+$18.4
& \textbf{73.4} & 66.2 & 72.0 & 85.6 & \textbf{82.5}
& 70.6 & \textbf{79.7} & 72.8 & 85.3 & 75.3 \\

Qwen2.5-VL 7B~\cite{qwen25vl2025}
& 76.2 & 72.8 & 79.0 & $-$6.2
& 69.0 & 72.6 & 74.3 & 91.1 & 78.3
& 71.0 & 78.4 & 71.3 & 87.5 & 77.9 \\
\bottomrule
\end{tabular}
}
\end{table*}

\begin{table*}[tb]
\caption{\textbf{Counterfactual weather augmentation and linguistic variation performance on the Adverse Weather task.} SCS measures the percentage of original--counterfactual image pairs for which the model changes its prediction. \textbf{Bold} and \underline{underlined} values denote the best and worst results.}
\label{tab:aw_axis23}
\centering
\scriptsize
\setlength{\tabcolsep}{4pt}
\renewcommand{\arraystretch}{1.13}
\resizebox{\textwidth}{!}{
\begin{tabular}{@{}lrrrrp{3.5cm}|rrp{3.8cm}@{}}
\toprule
& \multicolumn{5}{c|}{\textit{Counterfactual SCS (\%)}}
& \multicolumn{3}{c}{\textit{Linguistic Variation}} \\
\cmidrule(lr){2-6} \cmidrule(lr){7-9}
Model & Overall & Fog & Rain & Snow & Observation
& Para.\ (\%) & Neg.\ (\%) & Observation \\
\midrule
Moondream 2B~\cite{moondream2025}  & \underline{13.4} & \underline{16.7} & \underline{16.0} & \underline{11.6}
& Lowest SCS; answers largely weather-invariant
& 80.0 & \underline{36.6} & Stable under paraphrase; worst negation \\
 
PaliGemma 3B~\cite{paligemma2024}  & \textbf{39.3} & \textbf{39.6} & \textbf{43.1} & \textbf{36.3}
& Most sensitive; strongest response to rain
& 80.9 & 48.6 & High stability; near-chance negation \\
 
SmolVLM 2.2B~\cite{smolvlm2025} & 28.1 & 32.2 & 29.0 & 26.6
& Moderate; fog induces largest shift
& \underline{78.5} & 43.2 & Stable paraphrase; below-chance negation \\

LLaVA-OV 8B~\cite{llavaonevision2024}  & 32.6 & 37.3 & 36.3 & 32.0
& Strong; consistently reacts across all weather
& 80.0 & 44.2 & Robust to rewording; weak polarity handling \\

InternVL3 8B~\cite{internvl32025}  & 27.9 & 30.9 & 31.5 & 28.8
& Moderate; balanced across weather types
& \textbf{87.1} & 52.2 & Best paraphrase among 5B; only model above chance \\

Qwen2.5-VL 7B~\cite{qwen25vl2025} & 34.2 & 38.1 & 42.0 & 23.4
& Second-most sensitive overall; strongest fog response
& 86.8 & \textbf{73.4} & Only model clearly above chance on negation; large margin over next-best \\
\midrule
Mean  & 29.2 & 32.5 & 33.0 & 26.4
& & & & \\
\bottomrule
\end{tabular}
}
\end{table*}

\noindent\textbf{Semantic question types.} 
Performance also depends strongly on the reasoning required by the question. Scene-understanding questions are consistently the easiest across models, whereas hazard-related questions remain the most challenging, particularly for smaller VLMs. This gap suggests that recognizing global scene characteristics is substantially easier than making safety-critical decisions that require integrating localized visual cues with driving context. Consequently, high performance on scene-level perception should not be interpreted as evidence of reliable decision-making for autonomous driving (Fig.~\ref{fig:adv_weater_primary}).



\noindent\textbf{Counterfactual Scene-Change Sensitivity (SCS) and Linguistic Robustness.}
Table~\ref{tab:aw_axis23} evaluates whether VLMs adapt their decisions when either the visual scene or the linguistic formulation changes while preserving the underlying driving task. Counterfactual weather transformations probe visual grounding by measuring whether models revise their predictions when clear scenes are converted into fog, rain, or snow, whereas linguistic variants assess robustness to paraphrasing and semantic polarity reversal.


Counterfactual evaluation reveals that current VLMs are often insensitive to safety-relevant visual changes. PaliGemma~3B exhibits the highest SCS (39.3\%), indicating that its decisions respond most consistently to changing weather conditions, whereas Moondream~2B remains largely invariant (13.4\%), suggesting that its predictions are driven more by learned answer priors than by visual evidence. Across models, rain and fog induce larger response changes than snow, yet the overall SCS remains modest, indicating that even substantial degradations in visibility frequently fail to alter model decisions. For autonomous driving, such behavior is undesirable, as weather-dependent hazards should naturally lead to different driving actions (Fig.~\ref{fig:adv_weater_primary}).

 
Linguistic robustness exposes a complementary limitation. While all models remain highly consistent under paraphrasing (78.5--87.1\%), negation accuracy collapses to near-chance for five of the six evaluated models. Qwen2.5-VL~7B is the only clear exception, achieving 73.4\% negation accuracy, demonstrating that robust polarity reasoning is attainable but not yet common among current open-source VLMs. The large gap between paraphrase consistency and negation accuracy indicates that these models are not generally sensitive to wording changes; instead, they struggle specifically with semantic polarity reversal. In safety-critical driving, such failures can directly invert decisions, for example confusing whether a vehicle \textit{should} or \textit{should not} stop (Fig.~\ref{fig:adv_weater_primary}).


Together, these results expose two core limitations of current VLMs in autonomous driving: (i) limited visual grounding, where models fail to update decisions under meaningful scene changes, and (ii) weak polarity-aware reasoning, where subtle linguistic changes lead to incorrect driving decisions despite otherwise stable language understanding.

\begin{table*}[tb]
\caption{\textbf{Image-conditioned performance on the Junctions \& Intersections task.} Results are reported in accuracy. The affirmation gap (A. Gap) quantifies response bias, where positive (negative) values indicate over-affirmative (over-negative) behavior. \textbf{Bold} and \underline{underlined} values denote the best and worst results.}
\label{tab:ji_axis1}
\centering
\scriptsize
\setlength{\tabcolsep}{3pt}
\renewcommand{\arraystretch}{1.13}
\resizebox{\textwidth}{!}{%
\begin{tabular}{@{}lrrrr|rrrrr@{}}
\toprule
& \multicolumn{4}{c|}{Language-prior probing (\%)}
& \multicolumn{5}{c}{Semantic question types (\%)} \\
\cmidrule(lr){2-5} \cmidrule(lr){6-10}
Model
& Overall & GT=yes & GT=no & A. Gap
& Perc. & Action & Hazard & Scene & Adv. \\
\midrule
Moondream 2B~\cite{moondream2025}
& \underline{63.7} & \textbf{85.0} & \underline{45.5} & $+$39.5
& 82.1 & \underline{51.9} & \underline{34.0} & \underline{77.1} & \underline{37.9} \\
 
PaliGemma 3B~\cite{paligemma2024}
& 71.1 & 73.0 & 69.6 & $+$3.4
& \underline{78.3} & 61.6 & 38.1 & 82.0 & 69.7 \\
 
SmolVLM 2.2B~\cite{smolvlm2025}
& 73.5 & \underline{71.7} & 74.9 & $-$3.2
& 85.5 & 67.1 & 47.7 & 77.7 & 66.3 \\
 
LLaVA-OV 8B~\cite{llavaonevision2024}
& 73.5 & 75.4 & 72.0 & $+$3.4
& \textbf{87.0} & 57.1 & 52.7 & \textbf{88.9} & 56.7 \\
 

InternVL3 8B~\cite{internvl32025}
& \textbf{80.2} & 78.8 & \textbf{81.3} & $-$2.5
& 86.7 & 67.3 & \textbf{80.9} & 86.7 & 72.8 \\

Qwen2.5-VL 7B~\cite{qwen25vl2025}
& 78.9 & 77.5 & 80.0 & $-$2.5
& 87.0 & \textbf{71.0} & 56.6 & 88.7 & \textbf{76.4} \\
 \bottomrule
\end{tabular}
}
\end{table*}

\subsection{Performance Analysis on Junctions \& Intersections}

\noindent\textbf{Language-prior (with Image).}
Table~\ref{tab:ji_axis1} reports image-conditioned performance on the Junctions \& Intersections benchmark. Although InternVL3~8B achieves the highest overall accuracy (80.2\%), GT-conditioned analysis reveals that comparable accuracies often arise from markedly different response biases. Moondream~2B exhibits a strong affirmation bias (+39.5\,pp), indicating a tendency to answer \texttt{yes} regardless of the visual scene, whereas InternVL3~8B and Qwen2.5-VL~7B show the most balanced response distributions ($-$2.5\,pp). These results reinforce a consistent finding across both benchmark categories: aggregate accuracy alone is insufficient for assessing reliability, as it conceals systematic decision policies that can dominate model behavior in safety-critical settings.

\begin{figure*}[t]
  \centering
 \includegraphics[width=\textwidth]{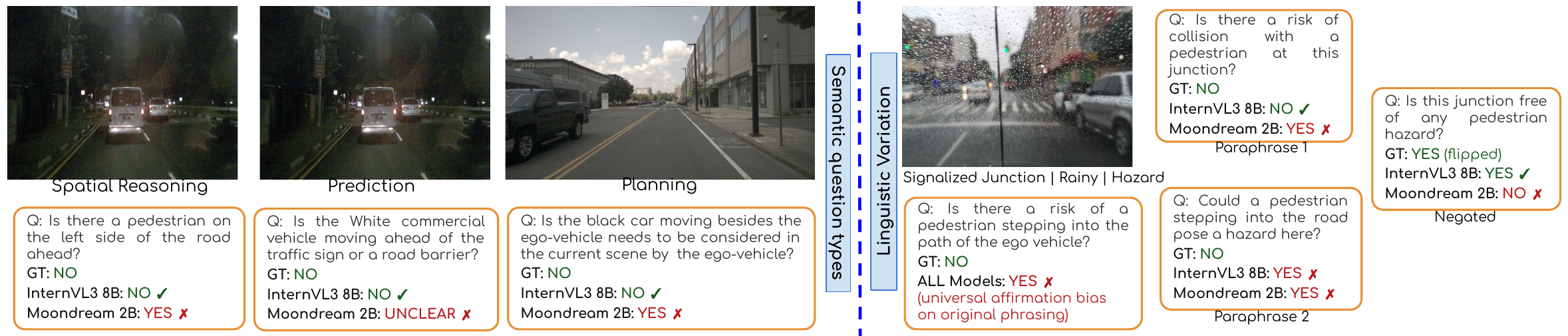}
  \caption{\textbf{Qualitative analysis of the Junctions \& Intersections task.} \textit{Semantic question types:} best- and worst-performing models on matched nuScenes examples spanning spatial reasoning, prediction, and planning. \textit{Linguistic variation:} responses to the original question, paraphrases, and the negated variant illustrate differing robustness to semantic reformulation, with InternVL3-8B partially recovering under paraphrasing and negation while Moondream-2B remains inconsistent. Zoom in for details; additional details are provided in the Supplementary.}
  \label{fig:jun_int_primary}
\vspace{-.5cm}
\end{figure*}

\noindent\textbf{Semantic question types.}
Performance across semantic question types exposes a more fundamental limitation. Perception and scene-understanding questions achieve consistently high accuracy (78--89\%) across all models, indicating that current VLMs can reliably recognize visible traffic elements and overall road context. In contrast, accuracy drops substantially on hazard and action questions, which require integrating multiple visual cues before making a driving decision. For example, Moondream and PaliGemma achieve only 34.0\% and 38.1\% on hazard questions, despite substantially stronger perception performance, while even larger models exhibit a noticeable gap between recognition and decision-making (Fig.~\ref{fig:jun_int_primary}).

This disparity suggests that current VLMs are considerably better at \emph{seeing} the scene than \emph{reasoning} about its driving implications. Recognizing pedestrians, traffic lights, or lane layouts does not reliably translate into judging whether they constitute an immediate hazard or determining an appropriate driving action. Qwen2.5-VL~7B further illustrates this disconnect, achieving the strongest action performance (71.0\%) yet trailing InternVL3~8B by more than 24\,pp on hazard reasoning (56.6\% vs.\ 80.9\%), demonstrating that competence in one driving skill does not generalize across related reasoning tasks.

Overall, the Junctions \& Intersections benchmark reveals that the primary limitation of current VLMs is not visual recognition, but the inability to consistently transform visual understanding into safe, context-aware driving decisions. This observation is further supported by the targeted reasoning analysis (Sec.~3.4), where many incorrect responses are accompanied by generic safety heuristics rather than evidence grounded in the observed traffic scene.

\noindent\textbf{nuScenes spatial reasoning and linguistic robustness.}
Table~\ref{tab:ji_axis23} evaluates whether the failure modes observed on BDD100K generalize to the DriveLM/nuScenes spatial reasoning benchmark. Although Qwen2.5-VL~7B achieves the highest overall accuracy (78.8\%), GT-conditioned analysis again reveals markedly different decision policies across models. Moondream~2B collapses to 36.2\% accuracy, exhibiting an extreme affirmation bias (92.3\% GT=yes vs.\ 14.2\% GT=no) together with a high unclear rate, whereas SmolVLM~2.2B achieves competitive overall performance by exhibiting the opposite tendency, strongly favoring \texttt{no} responses. These results reinforce that similar benchmark accuracies can arise from fundamentally different reasoning failures rather than genuinely balanced decision making.

Performance across question types further exposes the perception--reasoning gap. Spatial reasoning achieves consistently higher accuracy than planning across all models, suggesting that identifying object locations is considerably easier than translating those observations into prediction and driving decisions. Even on a benchmark designed for richer relational reasoning, current VLMs struggle to progress from recognizing scene geometry to reasoning about safe vehicle behavior.

Linguistic robustness exhibits the same trend observed in Adverse Weather. While paraphrase consistency remains moderate, negation accuracy is consistently near chance, dropping to 40.6\% for LLaVA-OV~8B. Even the best-performing model reaches only 56.8\%, indicating that failures under semantic polarity reversal persist across datasets and question types. Together, these results demonstrate that the limitations identified in BDD100K are not dataset-specific: current VLMs continue to rely on biased decision policies, exhibit limited reasoning beyond perception, and remain unreliable under linguistically challenging but safety-critical queries (Fig.~\ref{fig:jun_int_primary}).

\begin{table*}[tb]
\caption{\textbf{nuScenese spatial reasoning and linguistic variation performance on the Junctions \& Intersections task.} 
All accuracies exclude \texttt{unclear} responses. \textbf{Bold} and \underline{underlined} values denote the best and worst results.}
\label{tab:ji_axis23}
\centering
\scriptsize
\setlength{\tabcolsep}{3pt}
\renewcommand{\arraystretch}{1.13}
\resizebox{\textwidth}{!}{%
\begin{tabular}{@{}lrrrrrrp{3.2cm}|rrp{3.5cm}@{}}
\toprule
& \multicolumn{7}{c|}{\textit{nuScenes Spatial Reasoning (\%)}}
& \multicolumn{3}{c}{\textit{Linguistic Variation}} \\
\cmidrule(lr){2-8} \cmidrule(lr){9-11}
Model
& Overall & Spatial & Pred. & Plan.
& GT=yes & GT=no & Observation
& Para.\ (\%) & Neg.\ (\%) & Observation \\
\midrule
Moondream 2B~\cite{moondream2025}
& \underline{36.2} & \underline{48.2} & \underline{16.6} & \underline{51.1}
& \textbf{92.3} & \underline{14.2}
& Near-complete over-affirmation; high unclear rate
& \textbf{61.1} & 54.0
& Moderate paraphrase; above-chance negation \\
 
PaliGemma 3B~\cite{paligemma2024}
& 69.3 & 64.1 & 75.5 & 61.6
& 51.9 & 75.1
& Balanced; strong on prediction
& 57.6 & 49.4
& Low paraphrase stability; near-chance negation \\
 
SmolVLM 2.2B~\cite{smolvlm2025}
& 72.1 & 66.4 & 85.8 & 54.3
& \underline{41.3} & 83.5
& Extreme negation bias; defaults to \texttt{no}
& 56.2 & 50.3
& Low paraphrase stability; near-chance negation \\
 
LLaVA-OV 8B~\cite{llavaonevision2024}
& 71.8 & 70.8 & 76.3 & 60.7
& 62.4 & 74.9
& Balanced; consistent across q-types
& 57.8 & \underline{40.6}
& Moderate stability; worst negation \\
 

InternVL3 8B~\cite{internvl32025}
& 72.1 & 69.6 & 76.3 & \textbf{62.3}
& 67.1 & 73.2
& Balanced; ties for second-highest overall
& 57.9 & 54.0
& Best paraphrase; above-chance negation \\

Qwen2.5-VL 7B~\cite{qwen25vl2025}
& \textbf{78.8} & \textbf{75.4} & \textbf{87.5} & 60.2
& 51.3 & \textbf{88.0}
& Highest overall and strongest prediction accuracy, but weakest on GT=yes; leans toward default \texttt{no}
& \underline{55.3} & \textbf{56.8}
& Lowest paraphrase stability but highest negation accuracy across all six models \\
\bottomrule
\end{tabular}
}
\vspace{-.5cm}
\end{table*}




\subsection{Qualitative Analysis of Reasoning Failures}

To understand \emph{why} VLMs fail, we perform a targeted reasoning analysis on representative failure cases using evidence-based prompts (see Supplementary Material). Across both benchmark tasks, correct explanations consistently reference observable scene evidence, whereas incorrect explanations are dominated by generic safety heuristics and hallucinated scene attributes rather than image-grounded observations. The same behavior persists without images, where models often produce confident driving recommendations instead of acknowledging insufficient visual information. These qualitative findings explain the strong no-image performance observed quantitatively and confirm that many failures stem from reliance on language priors rather than visual grounding.

\begin{figure*}[tb]
  \centering
  \includegraphics[width=\textwidth]{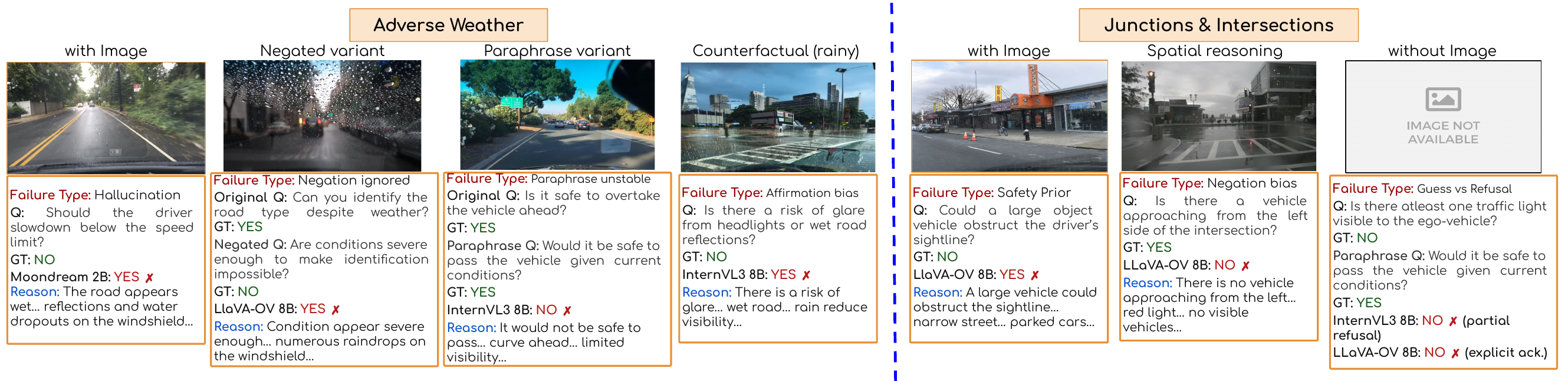}
  \caption{\textbf{Targeted qualitative analysis of reasoning failures.}
Representative failure cases from each diagnostic protocols showing that VLMs often generate plausible driving decisions from language priors rather than visual evidence. Examples illustrate hallucination, safety-prior reasoning, negation blindness, paraphrase inconsistency, counterfactual scene insensitivity, spatial reasoning failure, and no-image guess-versus-refusal behavior. Each panel reports the question, ground truth, prediction, and generated rationale. Zoom in for details.}
  \label{fig:phase2_failures}
\vspace{-.5cm}
\end{figure*}

Fig.~\ref{fig:phase2_failures} presents representative failure cases across all diagnostic evaluation protocols. Despite different tasks and prompting conditions, the same failure patterns consistently emerge. Under image-conditioned evaluation, models frequently substitute visual evidence with generic safety priors, hallucinating hazards (e.g., wet roads or sightline obstructions) that are absent from the scene. Linguistic perturbations further expose poor semantic robustness: negated questions are often interpreted identically to their affirmative counterparts, while meaning-preserving paraphrases can unexpectedly reverse model decisions. Counterfactual weather transformations reveal limited visual grounding, with models either failing to adapt their predictions when scene conditions change or exhibiting strong affirmation bias irrespective of the visual evidence. On the nuScenes extension, spatial reasoning failures arise from incorrect inference of object presence despite clearly observable scene geometry. Finally, in the absence of images, models either generate confident but unsupported driving recommendations or inconsistently refuse to answer, demonstrating that plausible responses can be produced without any visual evidence. Additional qualitative examples and detailed reasoning traces are provided in the Supplementary Material.

\section{Conclusion}
\label{sec:conclusion}

We presented a diagnostic evaluation framework for assessing whether vision-language models make safety-critical driving decisions from visual evidence or from language priors. Rather than relying on aggregate accuracy, our framework evaluates visual grounding through four evaluation protocols and metrics: image versus no-image evaluation, counterfactual weather sensitivity, linguistic robustness, and spatial reasoning. Across six representative open-source VLMs evaluated on BDD100K and a DriveLM/nuScenes spatial reasoning extension, we find that competitive benchmark accuracy consistently masks systematic failure modes.
Our analysis reveals strong reliance on language priors, limited sensitivity to counterfactual scene changes, persistent affirmation and negation biases, and confident decisions even without visual input. These findings suggest that current open-source VLMs remain insufficiently grounded for safety-critical autonomous driving despite producing plausible natural-language explanations.

We hope this benchmark encourages future research to move beyond aggregate accuracy toward evaluations that explicitly measure visual grounding, robustness, and reasoning reliability, ultimately enabling more trustworthy vision-language models for autonomous driving.
\bibliographystyle{splncs04}
\bibliography{main}
\end{document}